\pdfoutput=1
\documentclass[11pt]{article}

\usepackage{multirow}
\usepackage{graphicx}
\usepackage{booktabs}
\usepackage{colortbl}
\usepackage{paralist}
\usepackage{enumitem}
\usepackage{amssymb}
\usepackage{hyperref}
\usepackage{bm}
\usepackage{makecell}

\usepackage{ACL2023}

\usepackage{times}
\usepackage{latexsym}

\usepackage[T1]{fontenc}
\usepackage[utf8]{inputenc}

\usepackage{microtype}

\usepackage{inconsolata}

\title{\textit{Learn over Past, Evolve for Future:} \\Forecasting Temporal Trends for Fake News Detection}

\newcommand{\emailsI}{\href{mailto:hubeizhe21s@ict.ac.cn}{hubeizhe21s},\href{mailto:shengqiang18z@ict.ac.cn}{shengqiang18z},\href{mailto:caojuan@ict.ac.cn}{caojuan},\href{mailto:zhuyongchun18s@ict.ac.cn}{zhuyongchun18s}}
\newcommand{\emailsII}{\href{mailto:wangdanding@ict.ac.cn}{wangdanding},\href{mailto:wangzhengjia21b@ict.ac.cn}{wangzhengjia21b}}

\author{
Beizhe Hu$^{1,2}$\quad
Qiang Sheng$^{1,2,}$\thanks{$^*$Corresponding author.}\quad
Juan Cao$^{1,2}$\quad
Yongchun Zhu$^{1,2}$ \\
{\bf Danding Wang}$^{1}$ \quad
{\bf Zhengjia Wang}$^{1,2}$ \quad
{\bf Zhiwei Jin}$^{3}$ \\
	$^{1}$Key Lab of Intelligent Information Processing of Chinese Academy of Sciences, \\
	Institute of Computing Technology, Chinese Academy of Sciences\\
	$^{2}$University of Chinese Academy of Sciences \quad
    $^{3}$ ZhongKeRuijian Technology Co., Ltd.\\
	\texttt{\{\emailsI\}@ict.ac.cn}\\
	\texttt{\{\emailsII\}@ict.ac.cn}, \texttt{\href{mailto:jinzhiwei@ruijianai.com}{jinzhiwei}@ruijianai.com}
}

\begin{document}
\maketitle
\begin{abstract}
Fake news detection has been a critical task for maintaining the health of the online news ecosystem. However, very few existing works consider the temporal shift issue caused by the rapidly-evolving nature of news data in practice, resulting in significant performance degradation when training on past data and testing on future data. In this paper, we observe that the appearances of news events on the same topic may display discernible patterns over time, and posit that such patterns can assist in selecting training instances that could make the model adapt better to future data. Specifically, we design an effective framework \textbf{FTT} (\textbf{F}orecasting \textbf{T}emporal \textbf{T}rends), which could forecast the temporal distribution patterns of news data and then guide the detector to fast adapt to future distribution.
Experiments on the real-world temporally split dataset demonstrate the superiority of our proposed framework. The code is available at \href{https://github.com/ICTMCG/FTT-ACL23}{https://github.com/ICTMCG/FTT-ACL23}.
\end{abstract}

\section{Introduction}
Automatic fake news detection, which aims at distinguishing inaccurate and intentionally misleading news items from others automatically, has been a critical task for maintaining the health of the online news ecosystem~\cite{shu2017survey}. As a complement to manual verification, automatic fake news detection enables efficient filtering of fake news items from a vast news pool. Such a technique has been employed by social media platforms like Twitter to remove COVID-19-related misleading information during the pandemic~\cite{roth2022tweet}.

\begin{figure}[t]
\setlength{\belowcaptionskip}{-0.5cm}
	\centering
	\includegraphics[width=\linewidth]{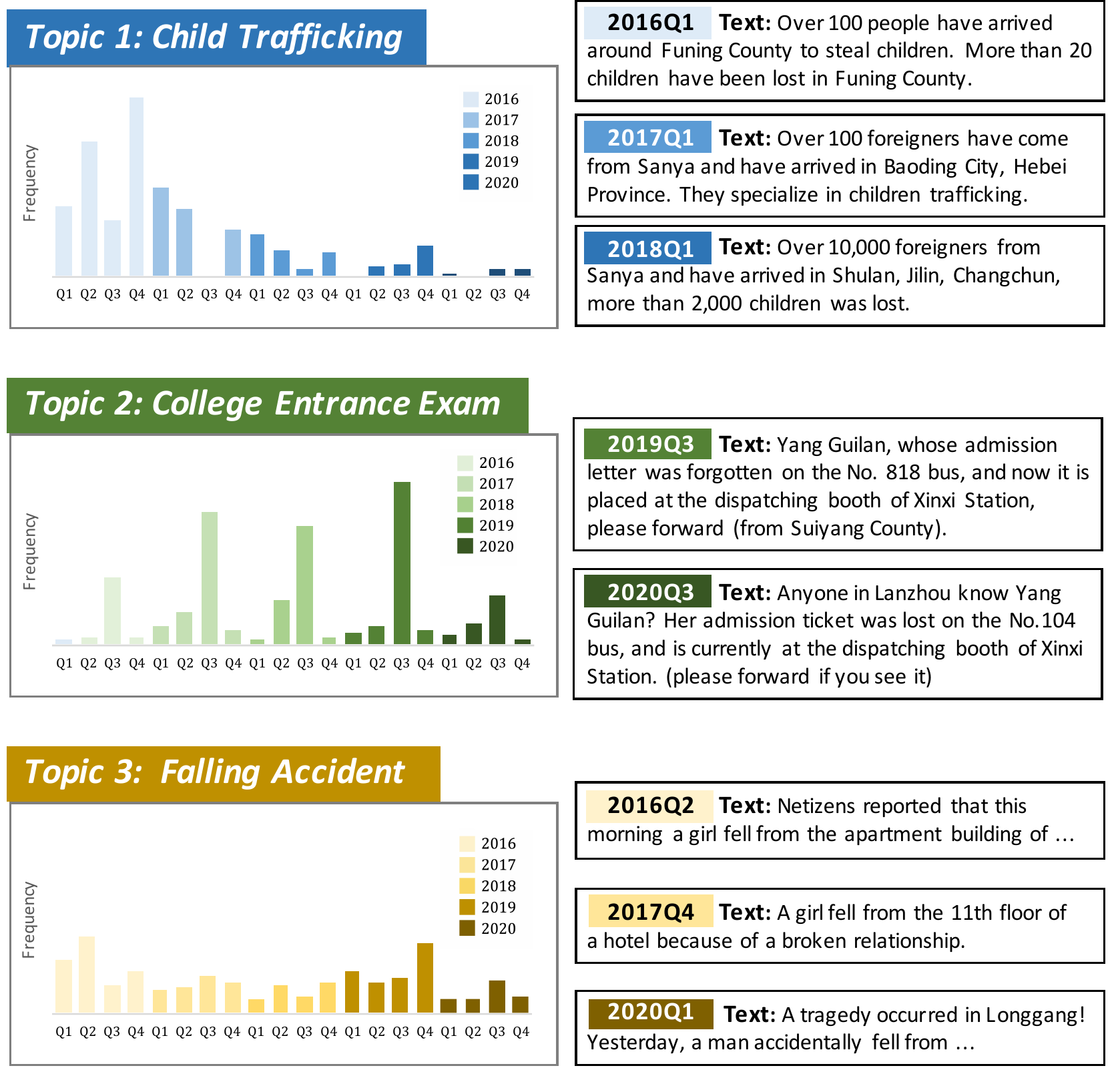}
	\caption{Topic-level statistics of news items across five years in our data. We see that different topics present diverse temporal patterns such as \textit{decrease} (Topic 1), \textit{periodicity} (Topic 2), and \textit{approximate stationery} (Topic 3), which we rely on to forecast temporal trends for better fake news detection in the future. The case texts are translated from Chinese into English.}
	\label{fig:eg}
\end{figure}

Over the past decade, most fake news detection researchers have followed a conventional paradigm of collecting a fixed dataset and \textit{randomly} dividing it into training and testing sets. However, the assumption that news data subsets are independent and identically distributed often does not hold true in real-world scenarios.
In practice, a fake news detection model is trained on \textit{offline data} collected up until the current time period but is required to detect fake news in newly arrived \textit{online data} at the upcoming time period.
Due to the rapidly-evolving nature of news, news distribution can vary with time, namely \textit{temporal shift}~\cite{du2021adarnn,gaspers-etal-2022-temporal}, leading to the distributional difference between offline and online data.
Recent empirical studies~\cite{dual-emotion,mu-etal-2023-time} evidence that fake news detection models suffer significant performance degradation when the dataset is temporally split. Therefore, the temporal shift issue has been a crucial obstacle to real-world fake news detection systems.

The temporal shift scenario presents a more significant challenge than common \textit{domain shift} scenarios.
Most existing works on the domain shift in fake news detection focus on transfer among pre-defined news channels (e.g., politics)~\cite{silva2021embracing,shukai-www22,lin-etal-2022-detect,nan-coling}. However, consecutive data slices over time have various types of temporal dependencies and non-explicit distributional boundaries, making the temporal shift challenging. Moreover, these works assume the availability of target domain data, which is impossible for the temporal shift scenarios. Under such constraints,  our aim is to train a model using presently available data to generalize to future online data (corresponding to temporal generalization task; \citealp{dg-survey}).
Others that improve the generalizability to unseen domains learn domain-invariant features by adversarial learning~\cite{eann} and domain-specific causal effect removal~\cite{endef}, but do not consider the characteristics of temporal patterns of news events.

In this paper, we posit that the appearance of news events on the same topic presents diverse temporal patterns, which can assist in evaluating the importance of previous news items and boost the detection of fake news in the upcoming time period.
In \figurename~\ref{fig:eg}, we exemplify this assumption using the statistics of news items on three topics in the Chinese Weibo dataset:
Topic 1 presents the temporal pattern of \textit{decrease}, where news about child trafficking becomes less frequent. 
Topic 2 presents the \textit{periodicity} of news related to the college entrance exam which takes place annually in the second quarter (Q2).\footnote{We denote the four quarters of a calendar year as Q1-Q4, respectively. For instance, Q1 stands for January through March.} 
In Topic 3, news items about falling accidents appear repeatedly and exhibit an \textit{approximate stationary} pattern.
Such temporal patterns indicate the different importance of news samples in the training set for detection in future quarters. For instance, instances of Topic 2 in the training set are particularly important for effectively training the detector to identify fake news in Q3.

\begin{figure*}[htbp]
\setlength{\belowcaptionskip}{-0.2cm}
	\centering
	\includegraphics[width=\textwidth]{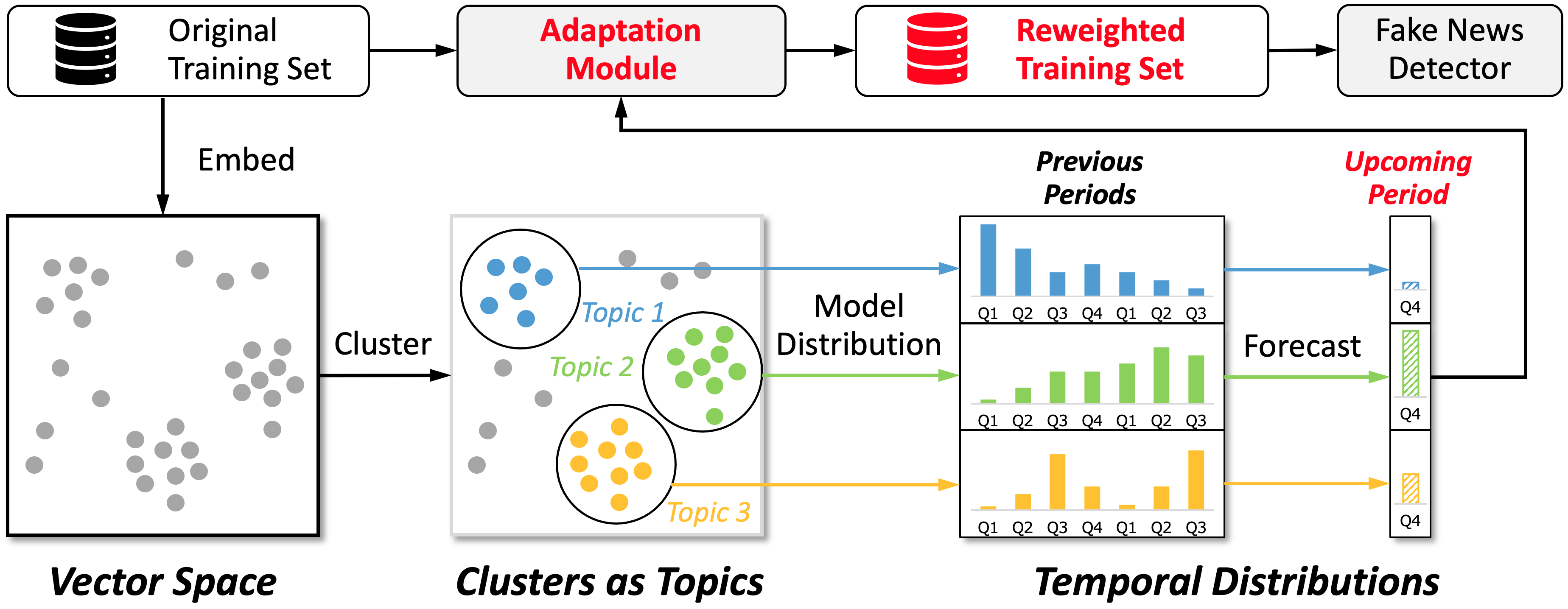}
	\caption{Architecture of the proposed \textbf{FTT} (\textbf{F}orecasting \textbf{T}emporal \textbf{T}rends) framework.
	}
	\label{fig:arch}
\end{figure*}

To this end, we propose to model the temporal distribution patterns and forecast the topic-wise distribution in the upcoming time period for better temporal generalization in fake news detection, where the forecasted result guides the detector to fast adapt to future distribution.
\figurename~\ref{fig:arch} illustrates our framework \textbf{FTT} (\textbf{F}orecasting \textbf{T}emporal \textbf{T}rends). We first map training data to vector space and perform clustering to discover topics. Then we model the temporal distribution and forecast the frequency of news items for each topic using a decomposable time series model. Based on the forecasts, we evaluate the importance of each item in the training data for the next time period by manipulating its weight in training loss. Our contributions are summarized as follows: 

\begin{compactitem}
    \item \textbf{Problem:} To the best of our knowledge, we are the first to incorporate the characteristics of topic-level temporal patterns for fake news detection.
    \item \textbf{Method:} We propose a framework for \textbf{F}orecasting \textbf{T}emporal \textbf{T}rends (\textbf{FTT}) to tackle temporal generalization issue in fake news detection.
    \item \textbf{Industrial Value:} We experimentally show that our FTT overall outperforms five compared methods while maintaining good compatibility with any neural network-based fake news detector.
\end{compactitem}

\section{Related Work}
\paragraph{Fake News Detection.} 
Fake news detection is generally formulated as a binary classification task between real and fake news items. Research on this task could be roughly grouped into content-only and social context-based methods. Content-only methods take the news content as the input including texts~\cite{sheng2021integrating}, images~\cite{qi2019exploiting}, and videos~\cite{bu2023combating}, and aim at finding common patterns in news appearances.
In this paper, we focus on textual contents but our method could be generalized to other modalities.
Previous text-based studies focus on sentiment and emotion~\cite{ajao,fakeflow}, writing style~\cite{style-aaai}, discourse structure~\cite{hdsf}, etc. Recent studies address the domain shift issues across news channels and propose multi-domain~\cite{mdfend,zhu2022memory} and cross-domain~\cite{nan-coling,lin-etal-2022-detect} detection methods. 
\citet{endef} design a causal learning framework to remove the non-generalizable entity signals.
Social context-based methods leverage crowd feedbacks~\cite{allinone,defend,dual-emotion}, propagation patterns~\cite{network-zhou,Propagation2Vec}, and social networks~\cite{fang,dc}, which have to wait for the accumulation of such social contexts.

Considering the in-time detection requirement, our proposed framework falls into the category of content-only methods, where we provide a new perspective for addressing the temporal generalization issue by forecasting temporal trends.

\paragraph{Temporal Generalization.}
The temporal generalization issue presents a situation in that models are trained on past data but required to perform well on unavailable and distribution-shifted future data. It has been addressed in a variety of applications such as review classification~\cite{huang-paul-2019-neural-temporality}, named entity recognition~\cite{temporal-ner}, and air quality prediction~\cite{du2021adarnn}. Recently, \citet{gaspers-etal-2022-temporal} explore several time-aware heuristic-based instance reweighting methods based on recency and seasonality for an industrial speech language understanding scenario. Our work follows this line of instance reweighting, but we attempt to model the temporal patterns and forecast topic-wise distribution to better adapt to future data.

\section{Proposed Framework}
Our framework FTT is presented in \figurename~\ref{fig:arch}, where the instances from past consecutive time periods in the original training set are reweighted according to the forecasted topic-wise distribution for generalizing better in the upcoming time period. In the following, we first provide the problem formulation and subsequently, detail the procedures.

\subsection{Problem Formulation}
Given a dataset $\mathcal{D}=\{\mathcal{D}_q\}_{q=1}^Q$ consisting of $Q$ subsets that contain news items from $Q$ consecutive time periods, respectively, our goal is to train a model on $\{\mathcal{D}_q\}_{q=1}^{Q-1}$ that generalizes well on $\mathcal{D}_Q$. In $\mathcal{D}$, an instance is denoted as $(x_i, y_i)$ where the ground-truth label $y_i=1$ if the content $x_i$ is fake.

In practice, we retrain and redeploy the fake news detector at a fixed time interval to reflect the effects of the latest labeled data. We set the interval as three months (i.e., a quarter) since a shorter interval does not allow sufficient accumulation of newly labeled fake news items. In the following, we set $\mathcal{D}_q$ as the subset corresponding to news in a quarter of a calendar year. 

\subsection{Step 1: News Representation}
We first transform the news content into a vector space to obtain its representation, which will be used for similarity calculation in the subsequent clustering step.
We employ Sentence-BERT~\cite{reimers2019sentence}, which is widely used for sentence representation (e.g.,\citealp{shaar}). For instance $\mathrm{x}_i$, the representation vector is $\bm{x}_i \in \mathbb{R}^{768}$.

\subsection{Step 2: Topic Discovery}
We perform clustering on news items based on the representation obtained in Step 1 to group news items into distinct clusters which correspond to topics.
Due to the lack of prior knowledge about the topic number, we adopt the single-pass incremental clustering algorithm which does not require a preset cluster number.
We first empirically set a similarity threshold $\theta_{sim}$ to determine when to add a new cluster.
When an item arrives, it is assigned to the existing cluster whose center is the nearest to it if the distance measured by cosine similarity is larger than $\theta_{sim}$. Otherwise, it will be considered as an item on a new topic and thus be in a new independent cluster.

\subsection{Step 3: Temporal Distribution Modeling and Forecasting}
Based on the clustering results, we model the temporal distribution of different news topics and forecast the topic-wise distribution in the upcoming time period in this step.
Note that we do not consider the clusters with news items less than the threshold $\theta_{count}$ since they are too small to present significant temporal patterns. 

\paragraph{Modeling.} 
Assuming that $T$ topics are preserved, we first count the number of news items per quarter within each topic. The counts of the same quarter are then normalized across topics to obtain the quarterly frequency sequence of each topic (denoted as $f$). To model the temporal distribution, we adopt a decomposable time series model~\cite{harvey1990estimation} on the quarterly sequences and consider the following two trends (exemplified using Topic $i$):

\textit{1) General Trend.} A topic may increase, decrease, or have a small fluctuation in terms of a general non-periodic trend (e.g., Topics 1 and 3 in~\figurename~\ref{fig:eg}). To fit the data points, we use a piecewise linear function:
\begin{equation}
        g_i(f_{i,q})=k_i f_{i,q}+m_i,
\end{equation}
where $k_i=k+\bm{a}(q)^T\bm{\delta}$ is the growth rate, $f_{i,q}$ is the frequency of Topic $i$ in Quarter $q$, and $m_i=m+\bm{a}(q)^T\bm{\gamma}$ is the offset. $k$ and $m$ are initial parameters. $\bm{a}(q)$ records the changepoints of growth rates and offsets while $\bm{\delta}$ is the rate adjustment term and $\bm{\gamma}$ is a smoothing term.

\textit{2) Quarterly Trend.}
For topics having quarterly periodic trends like Topic 2 in~\figurename~\ref{fig:eg}, we add four extra binary regressors corresponding to Q1\textasciitilde Q4 to inform the regression model the quarter that a data point in input sequence belongs to. For Topic $i$ and Quarter $q$, we obtain the quarterly seasonality function $s_i(f_{i,q})$ by summing the four regression models.

\paragraph{Forecasting.} 
We fit the model using the time series forecasting tool Prophet~\cite{prophet} with the temporal distribution of topics from Quarter 1 to Quarter $Q$-1. To forecast the trend of Topic $i$ in the upcoming Quarter $Q$, we sum up the two trend modeling functions:
\begin{equation}
    p_i(f_{i,Q}) = g_i(f_{i,Q}) + s_i(f_{i,Q}).
\end{equation}

\subsection{Step 4: Forecast-Based Adaptation}
Based on the topic-wise forecasts of frequency distribution in Quarter $Q$, we apply instance reweighting to the training set and expect the model trained using the reweighted set would better adapt to the future data in Quarter $Q$.

We first filter out topics that do not exhibit obvious regularity. Specifically, we remove the topics which have a mean absolute percentage error (MAPE) larger than a threshold $\theta_{mape}$ during the regression fitting process.
For a Topic $i$ in the preserved set $\mathcal{D_Q}^\prime$, we calculate and then normalize the ratio between the forecasted frequency of Topic $i$ $p_i(f_{i,Q})$ and the sum of all forecasted frequencies of the preserved topics:
\begin{equation}
    w_{i,Q} = \mathrm{Bound}\left(\frac{p_i(f_{i,Q})}{\sum_{i \in {D_Q}^\prime} p_i(f_{i,Q})}\right),
\end{equation}
where $\mathrm{Bound}$ is a function to constrain the range of calculated weights. We set the weight smaller than $\theta_{lower}$ and larger than $\theta_{upper}$ as $\theta_{lower}$ and $\theta_{upper}$, respectively, to avoid the instability during the training process. For those that are not included in $\mathcal{D_Q}^\prime$, we set their weights as 1.

The new weight of the training set instances of Topic $i$, $w_{i,Q}$, corresponds to our forecasts of how frequent news items of this topic will emerge in the upcoming period $Q$.
If the forecasted frequency of Topic $i$ indicates a decreasing trend, the value will be smaller than 1 and thus instances of this topic will be down-weighted; conversely, if the forecasted distribution indicates an increasing trend, the value will be greater than 1 and the instances will be up-weighted. In the next step, we will show the reweighting process during training.

\subsection{Step 5: Fake News Detector Training}
Our framework FTT could be compatible with any neural network-based fake news detector. Here, we exemplify how FTT helps detectors' training using a pretrained BERT model~\cite{bert}.
Specifically,  given an instance $\bm{x}_i$, we concatenate the special token $\mathtt{[CLS]}$ and $\bm{x}_i$, and feed them into BERT. 
The average output representation of non-padded tokens, denoted as $\bm{o}_i$, is then fed into a multi-layer perception ($\mathrm{MLP}$) with a $\mathrm{sigmoid}$ activation function for final prediction:
\begin{equation}
    \hat{y}_i = \mathrm{sigmoid}(\mathrm{MLP}(\bm{o}_i)).
\end{equation}
Our difference lies in using the new weights based on the forecasted temporal distribution to increase or decrease the impact of instances during back-propagation. Unlike most cases that use an \textit{average} cross-entropy loss, we minimize the \textit{weighted} cross-entropy loss function during training:
\begin{equation}
    \mathcal{L} = -\frac{1}{N}\sum_{i=1}^N w_{i,Q} \mathrm{CrossEntropy}(y_i,\hat{y}_i),
\end{equation}
where $w_{i,Q}$ is the new weight for instance $x_i$ and $y_i$ is its ground-truth label. $N$ is the size of a mini-batch of the training set.

\begin{table*}[htbp]
  \centering
  \resizebox{0.95\linewidth}{!}{
    \setlength{\tabcolsep}{1.5mm}{
    \begin{tabular}{c | p{2.0cm}<{\centering} | p{2.18cm}<{\centering} p{2.18cm}<{\centering} p{2.18cm}<{\centering} p{2.18cm}<{\centering} p{2.18cm}<{\centering} p{2.18cm}<{\centering}}
    \toprule
     \textbf{2020} & \textbf{Metric} & \textbf{Baseline} & \textbf{EANN$_{T}$} &\textbf{ \makecell[c]{Same Period\\Reweighting} } & \textbf{\makecell[c]{Prev. Period \\Reweighting}} & \textbf{\makecell[c]{Combined\\Reweighting}} & \textbf{FTT (Ours)}\\
    \midrule
    \multirow{4}{*}{\textbf{Q1}} & macF1 & 0.8344 & 0.8334 & 0.8297 & 0.8355 & 0.8312 & \textbf{0.8402}\\
    & Accuracy & 0.8348 & 0.8348 & 0.8301 & 0.8359 & 0.8315 & \textbf{0.8409}\\
    & F1$_\mathrm{fake}$ & 0.8262 & 0.8181 & 0.8218 & 0.8274 & 0.8237 & \textbf{0.8295}\\
    & F1$_\mathrm{real}$ & 0.8425 & 0.8487 & 0.8377 & 0.8435 & 0.8387 & \textbf{0.8509}\\
    \midrule
    \multirow{4}{*}{\textbf{Q2}} & macF1 & 0.8940 & 0.8932 & 0.8900 & 0.9004 & 0.8964 & \textbf{0.9013}\\
    & Accuracy & 0.8942 & 0.8934 & 0.8902 & 0.9006 & 0.8966 & \textbf{0.9014}\\
    & F1$_\mathrm{fake}$ & 0.8894 & 0.8887 & 0.8852 & 0.8953 & 0.8915 & \textbf{0.8981}\\
    & F1$_\mathrm{real}$ & 0.8986 & 0.8978 & 0.8949 & \textbf{0.9055} & 0.9013 & 0.9046\\
    \midrule
    \multirow{4}{*}{\textbf{Q3}} & macF1 & 0.8771 & 0.8699 & 0.8753 & 0.8734 & 0.8697 & \textbf{0.8821}\\
    & Accuracy & 0.8776 & 0.8707 & 0.8759 & 0.8741 & 0.8707 & \textbf{0.8827}\\
    & F1$_\mathrm{fake}$ & 0.8696 & 0.8593 & 0.8670 & 0.8640 & 0.8582 & \textbf{0.8743}\\
    & F1$_\mathrm{real}$ & 0.8846 & 0.8805 & 0.8836 & 0.8829 & 0.8812 & \textbf{0.8900}\\
    \midrule
    \multirow{4}{*}{\textbf{Q4}} & macF1 & 0.8464 & 0.8646 & 0.8464 & 0.8429 & 0.8412 & \textbf{0.8780}\\
    & Accuracy & 0.8476 & 0.8647 & 0.8476 & 0.8442 & 0.8425 & \textbf{0.8784}\\
    & F1$_\mathrm{fake}$ & 0.8330 & 0.8602 & 0.8330 & 0.8286 & 0.8271 & \textbf{0.8707}\\
    & F1$_\mathrm{real}$ & 0.8598 & 0.8690 & 0.8598 & 0.8571 & 0.8553 & \textbf{0.8853}\\
    \midrule
    \multirow{4}{*}{\textbf{Average}} & macF1 & 0.8630 & 0.8653 & 0.8604 & 0.8631 & 0.8596 & \textbf{0.8754}\\
    & Accuracy & 0.8636 & 0.8659 & 0.8610 & 0.8637 & 0.8603 & \textbf{0.8759}\\
    & F1$_\mathrm{fake}$ & 0.8546 & 0.8566 & 0.8518 & 0.8538 & 0.8501 & \textbf{0.8682}\\
    & F1$_\mathrm{real}$ & 0.8714 & 0.8740 & 0.8690 & 0.8723 & 0.8691 & \textbf{0.8827}\\
    \bottomrule
    \end{tabular}
    }
  }
  \caption{Performance of the baseline method, four existing methods, and our method in fake news detection. The best result in each line is \textbf{bolded}.}
  \label{tab:main_result}%
\end{table*}%

\section{Evaluation}
    \label{sec:exp}
    We conduct experiments to answer the following evaluation questions:
    \begin{compactitem}
     \item \textbf{EQ1:} 
     Can FTT bring improvement to the fake news detection model in temporal generalization scenarios?
     \item \textbf{EQ2:} How does FTT help with fake news detection models?
    \end{compactitem}

\subsection{Dataset}
Our data comes from a large-scale Chinese fake news detection system, covering the time period from January 2016 to December 2020. To meet the practical requirements, the data was divided by quarters based on the timestamp. Unlike the existing academic datasets~\cite{fakenewsnet,sheng-etal-2022-zoom}, the dataset is severely imbalanced. To avoid instability during training, we randomly undersampled the subset of each quarter to achieve a ratio of 1:1 between fake and real news.
Identical to the real-world setting, we adopt a \textit{rolling training} experimental setup. If we train a model to generalize well in the time period $Q$, the training, validation, and testing sets would be $\{\mathcal{D}_i\}_{i=1}^{Q-2}$, $\mathcal{D}_{Q-1}$, and $\mathcal{D}_Q$, respectively. If the target is $Q+1$, then the three subsets would be $\{\mathcal{D}_i\}_{i=1}^{Q-1}$, $\mathcal{D}_{Q}$, and $\mathcal{D}_{Q+1}$.
Here we use the four quarterly datasets from 2020 as the testing sets and conduct experiments on the four sets separately.

\subsection{Experimental Settings}
\paragraph{Compared Methods.}
We compared our proposed FTT with five existing methods (including the vanilla baseline model), in which the second one is to remove non-generalizable bias and the last three are to introduce heuristic rules for adapting to future data.

\begin{itemize}
     \item \textbf{Baseline} follows a normal training strategy where all training instances are equally weighted.
     \item \textbf{EANN$_{T}$}~\cite{eann} is a model that enhances model generalization across events by introducing an auxiliary adversarial training task to prevent the model from learning event-related features. For fair comparison, we replaced the original TextCNN~\cite{textcnn} with a trainable BERT as the  textual feature extractor, and utilized publication year labels as the labels for the auxiliary task following \citet{endef}. We removed the image branch in EANN as here we focus on text-based fake news detection.
     \item \textbf{Same Period Reweighting} increases the weights of all training instances from the same quarter as the target data. It models the seasonality in the time series data.
     \item \textbf{Previous Period Reweighting} increases the weights of all training instances from the last quarter. It could capture the recency in the data distribution.
     \item \textbf{Combined Reweighting} combines the two reweighting methods mentioned above. The last three methods are derived from \cite{gaspers-etal-2022-temporal}.
\end{itemize}

\paragraph{Implementation Details.}
We used a BERT model, \texttt{hfl/chinese-bert-wwm-ext}~\cite{chinese-bert} implemented in HuggingFace's Transformer Package~\cite{transformer} as the baseline fake news detection classifier. In the training process, we used the Adam optimizer~\cite{adam} with a learning rate of 2e-5 and adopted the early stop training strategy, and reported the testing performance of the best-performing model on the validation set.
We employed grid search to find the optimal hyperparameters in each quarter for all methods. In Q1 and Q2, the optimal hyperparameters of FTT are $\theta_{sim}=0.65$, $\theta_{count}=30$, $\theta_{mape}=0.8$, $\theta_{lower}=0.3$, and $\theta_{upper}=2.0$; and in Q3 and Q4, they are $\theta_{sim}=0.5$, $\theta_{count}=30$, $\theta_{mape}=2.0$, $\theta_{lower}=0.3$, and $\theta_{upper}=2.0$.

We report the accuracy, macro F1 (macF1), and the F1 score for real and fake classes (F1$_\mathrm{real}$ and F1$_\mathrm{fake}$).

\subsection{Performance Comparison (EQ1)}
\tablename~\ref{tab:main_result} shows the overall and quarterly performance of the proposed framework and other methods. We observe that:

\textbf{1)} FTT outperforms the baseline and four other methods across all quarters in terms of most of the metrics (the only exception is F1$_\mathrm{real}$ in Q2). These results demonstrate its effectiveness.

\textbf{2)} The average improvement of F1$_\mathrm{fake}$ is larger than that of F1$_\mathrm{real}$, suggesting that our method helps more in capturing the uniqueness of fake news. We attribute this to the differences in temporal distribution fluctuation: fake news often focuses on specific topics, while real news generally covers more diverse ones. This makes the topic distribution of fake news more stable, which allows for better modeling of topic-wise distributions.

\textbf{3)} The three compared reweighting methods show inconsistent performances. In some situations, the performance is even lower than the baseline (e.g., Same Period Reweighting in Q1). We speculate that the failure is caused by the complexity of the news data. Considering the rapidly-evolving nature of news, single heuristic methods like recency and seasonality could not fast adapt to future news distribution. In contrast, our FTT performs topic-wise temporal distribution modeling and next-period forecasting and thus has a better adaption ability.

\begin{table}[htbp]
  \centering
  \setlength{\belowcaptionskip}{-0.5cm}\
  \resizebox{\linewidth}{!}{
    \setlength{\tabcolsep}{1.5mm}{
    \begin{tabular}{c | c | cc} 
    \toprule
     \textbf{Subset of the test set} & \textbf{Metric} & \textbf{Baseline}
     & \textbf{FTT (Ours)}\\
    \midrule
    \multirow{4}{*}{\textbf{Existing Topics}} & macF1 & 0.8425 & \textbf{0.8658}\\
    & Accuracy & 0.8589 & \textbf{0.8805}\\
    & F1$_\mathrm{fake}$ & 0.7997 & \textbf{0.8293}\\
    & F1$_\mathrm{real}$ & 0.8854 & \textbf{0.9023}\\

    % 单行 Out cluster
    \midrule
    % \rowcolor{gray!20}
    \multirow{4}{*}{\textbf{New Topics}} & macF1 & 0.8728 & \textbf{0.8846}\\
    % \rowcolor{gray!20}
    & Accuracy & 0.8729 & \textbf{0.8846}\\
    % \rowcolor{gray!20}
    & F1$_\mathrm{fake}$ & 0.8730 & \textbf{0.8849}\\
    % \rowcolor{gray!20} 
    & F1$_\mathrm{real}$ & 0.8727 & \textbf{0.8843}\\
    \bottomrule
    \end{tabular}%
    }
  }
  \caption{Breakdown of the performance on the testing set according to the existence of their topics.}
  \label{tab:ablation_study2}%
\end{table}%
\begin{figure}[t]
\setlength{\belowcaptionskip}{-0.3cm}
	\centering
	\includegraphics[width=\linewidth]{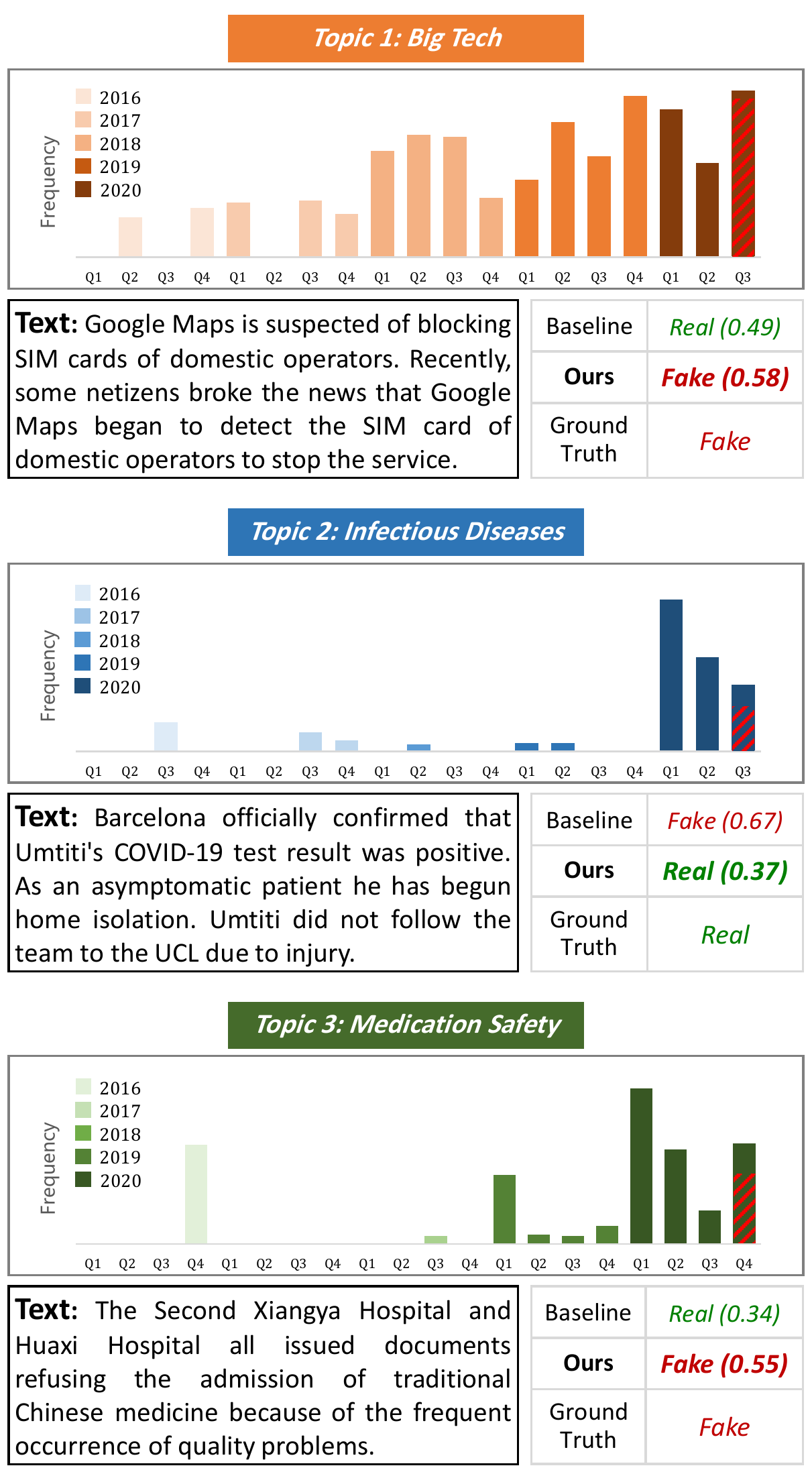}
	\caption{Three cases from the testing set. The forecasts by FTT about the frequency of the topics in the upcoming quarter are highlighted with {\color{red}red dashed bars}. The case texts are translated from Chinese into English.}
	\label{fig:case}
\end{figure}
\subsection{Result Analysis (EQ2)}
\paragraph{Statistical Analysis.} To analyze how FTT improves fake news detection performance, we analyze the testing instances by recognizing their topics. Specifically, we run the single-pass incremental clustering algorithm used in Step 2 again on the testing instances based on the clusters on the training set. If a news item in the testing set could be clustered into an existing cluster, it will be recognized as an item of the existing topics; otherwise, it will be in a new topic.
Based on the results, we show the breakdown of the performance on the testing set in \tablename~\ref{tab:ablation_study2}. Compared with the baseline, our framework achieves performance improvements on both the Existing Topics and the New Topics subsets. This could be attributed to our reweighting strategy where we not only increase the weights of news items belonging to a topic of an increasing trend but also decrease the weights of those belonging to the fading topics. With such a design, the model will be more familiar with news items in existing topics and more generalizable to news items in new topics.

\paragraph{Case Study.}
\figurename~\ref{fig:case} shows three cases from the testing set. According to the forecasted results of the frequencies of these topics in the testing time period, our framework assigns positive weights (greater than 1) to items in these topics. After training on the reweighted set, the detector flips its previously incorrect predictions. In Topic 1, the frequency of Big Tech-related news items demonstrated an increasing trend over time. FTT captures this pattern and provides a forecast close to the true value for the target quarter. In Topic 2, there is an explosive growth of Infectious Diseases-related news items in early 2020, followed by sustained high frequency in the subsequent quarters. FTT successfully captures this change. In contrast to the other two topics, the frequency of Medication Safety-related news items in Topic 3 exhibits both an overall increasing trend and a certain periodic pattern since 2019, which roughly follows a ``smiling curve'' from Q1 to Q4 in a single year. FTT effectively models both of these patterns and helps identify the importance of news items in this topic for the testing time period.

\section{Conclusion and Future Work}

We studied temporal generalization in fake news detection where a model is trained with previous news data but required to generalize well on the upcoming news data.
Based on the assumption that the appearance of news events on the same topic presents diverse temporal patterns, we designed a framework named FTT to capture such patterns and forecast the temporal trends at the topic level. The forecasts guided instance reweighting to improve the model's generalizability. Experiments demonstrated the superiority of our framework. In the future, we plan to mine more diverse temporal patterns to further improve fake news detection in real-world temporal scenarios.

\section*{Limitations}
We identify the following limitations in our work:

First, our FTT framework captures and models topic-level temporal patterns for forecasting temporal trends. Though the forecasts bring better temporal generalizability, FTT could hardly forecast the emergence of events in new topics. 

Second, FTT considers temporal patterns based on the topic-wise frequency sequences to identify patterns such as decrease,  periodicity, and approximate stationery. There might be diverse patterns that could not be reflected by frequency sequences.

Third, limited by the scarcity of the dataset that satisfies our evaluation requirements (consecutive time periods with a consistent data collection criterion), we only performed the experiments on a Chinese text-only dataset. Our method should be further examined on datasets of other languages and multi-modal ones.

\section*{Acknowledgements}
The authors thank anonymous reviewers for their insightful comments. This work was supported by the National Natural Science Foundation of China (62203425), the Zhejiang Provincial Key Research and Development Program of China (2021C01164), the Project of Chinese Academy of Sciences (E141020), and the Innovation Funding from Institute of Computing Technology, Chinese Academy of Sciences (E161020).

\bibliography{custom}
\bibliographystyle{acl_natbib}

\end{document}